\documentclass[letterpaper, conference]{IEEEtran}
\IEEEoverridecommandlockouts
\usepackage{textcomp}
\usepackage{xcolor}
\def\BibTeX{{\rm B\kern-.05em{\sc i\kern-.025em b}\kern-.08em
    T\kern-.1667em\lower.7ex\hbox{E}\kern-.125emX}}
\usepackage{geometry}
\usepackage{url}
\usepackage{comment}
\usepackage{amsmath,amssymb,amsfonts, mathtools}
\usepackage{graphicx}
\usepackage{bm}
\usepackage{epstopdf}
\usepackage{placeins}
\usepackage{bm}
\usepackage{cite}

\usepackage[inline]{enumitem}
\usepackage{multicol}
\usepackage{tikz}
\usepackage{caption}
\usepackage{subcaption}
\usepackage{tikz}
\usepackage{svg}
\usepackage{booktabs}
\usepackage{xcolor}
\usepackage{multirow}
\usepackage[linesnumbered,ruled,vlined]{algorithm2e}
\SetKwInput{KwInput}{Input}    % Set the Input
\SetKwInput{KwOutput}{Output}  % set the Output

\newcommand{\vect}[1]{\boldsymbol{#1}}
\newcommand\numberthis{\addtocounter{equation}{1}\tag{\theequation}}

\title{
\vspace{18pt} 
\LARGE \bf One-Shot Multimodal Learning from Demonstration with Force-Constrained Elastic Maps}

\author{Brendan Hertel, Jonathan Spanos, Navya Garg, and Reza Azadeh
\thanks{Authors are with the Persistent Autonomy and Robot Learning (PeARL) Lab, University of Massachusetts Lowell, Lowell, MA 01854, USA. Emails: \tt{\{brendan\_hertel, jonathan\_spanos, navya\_garg\}@student.uml.edu, reza@cs.uml.edu}}
}
 
\begin{document}

\maketitle
\thispagestyle{empty}
\pagestyle{empty}

\begin{abstract}

Robotic manipulation tasks often require simultaneous reasoning over motion and contact forces, yet most Learning from Demonstration (LfD) methods model only spatial trajectories and neglect force interactions with the environment. This limitation reduces robustness and can lead to unsafe or inconsistent task reproduction in force-constrained settings. We propose a novel one-shot multimodal LfD framework for the segmentation, encoding, and reproduction of force-inclusive demonstrations. First, we introduce a multimodal probabilistic segmentation method that adaptively weighs spatial and force modalities over time, enabling the automatic extraction of force-aware motion primitives. Second, we extend the elastic maps representation to incorporate external force constraints during skill encoding and formulate a convex optimization procedure for learning force-consistent trajectory models. The resulting skills reproduce both motion and contact characteristics from a single demonstration while promoting safer execution by accounting for demonstrated force profiles. We validate our approach on five real-world manipulation tasks across two distinct force-sensing configurations: wrist force sensing on a UR5e with a Robotiq 2f-85 gripper and finger force sensing on a Kinova Gen3 with an Openhand Model O gripper. Experimental results demonstrate robust multimodal segmentation, accurate force-aware reproduction, and cross-platform generality.

\end{abstract}

\section{Introduction}
\label{sec:intro}

Robots are continually increasing their ability to interact with the world around them. Just as humans rely on touch and feel to navigate and manipulate their environment, robot manipulators must be able to do the same. One of the primary ways robots can achieve this is through force sensing. A simple way of teaching robots new tasks is through Learning from Demonstration (LfD)~\cite{ravichandar2020recent}, where a human demonstrator shows a task, and the robot encodes and replicates the movement. Incorporating force data in these demonstrations provides robots with crucial context about the environment, allowing them to generate more informed and accurate movements during task reproduction. However, most LfD representations neglect forces, instead focusing solely on the robot’s spatial movements~\cite{Paraschos2013ProMP, Ahmadzadeh2018TLGC}. For robots to consistently and reliably perform tasks, force sensing must be integrated into LfD frameworks. 

\begin{figure}[t]
    \centering
    \includegraphics[width=0.98\linewidth]{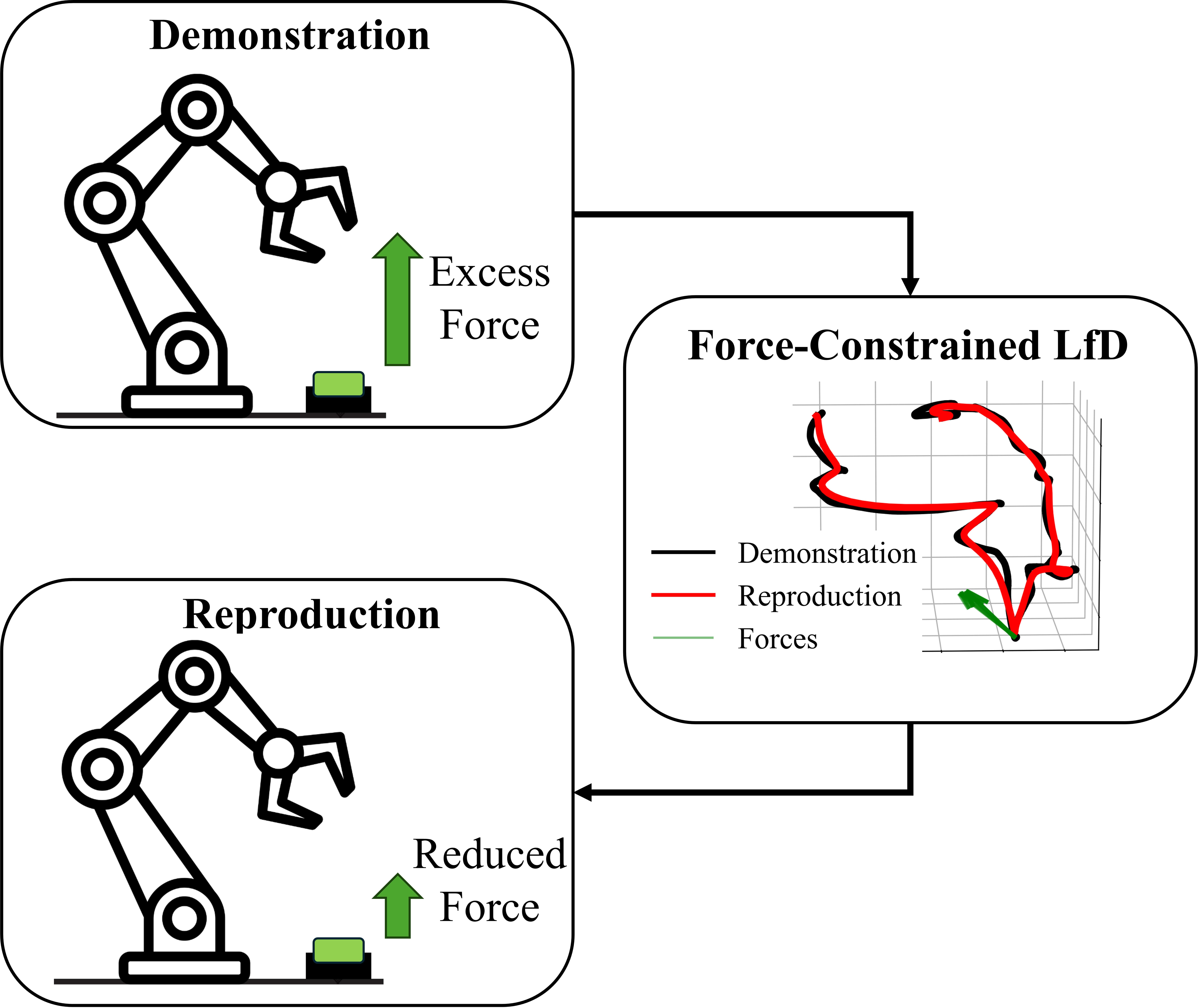}
    \caption{\small{Human demonstrations can sometimes include excessive forces that should not be replicated. Our force-constrained LfD framework avoids these forces resulting in more accurate and safer reproductions.}}
    \label{fig:fig1}
\end{figure}

In this paper, we propose a novel one-shot multimodal Learning from Demonstration (LfD) framework for the segmentation, encoding, and reproduction of force-constrained demonstrations. In this first phase, our framework employs a novel multimodal probabilistic segmentation technique to break down the demonstrations into a set of primitive movements. Segmentation is often done using only a single modality of a demonstration (usually position or vision)~\cite{nakamura2017segmenting}, and rarely with other modalities~\cite{hertel2024reusable_skills}. Our segmentation process, in contrast, can incorporate contextual information of the demonstrations from force data collected via force sensors. Data collection is straightforward due to the inclusion of force sensors located on either the wrist or fingertips of most modern robots. We validate our learning framework through experiments with both sensor configurations, showcasing its effectiveness in learning from force-constrained demonstrations. 

In the second phase, our framework encodes the segmented trajectories (i.e., primitives) using \emph{elastic maps}~\cite{hertel2022ElMap} to learn spatial trajectories, while uniquely integrating force constraints into the learning process. Notably, the extension of elastic maps to support external constraints is one of the novel contributions of this work. We then propose a convex formulation for optimizing the force-constrained elastic map model. The encoded models can be used to reproduce the learned skill, incorporating the forces from the demonstration into the reproduced motion, as seen in Fig.~\ref{fig:fig1}. Many robots have force/torque safety limits that can trigger during unsafe operation. A key benefit of our framework is its ability to preemptively avoid excessive forces by informing the skill learning process based on forces experienced during the demonstration.

We evaluate and validate the effectiveness of our approach on five real-world tasks, demonstrating its versatility across various force-sensing platforms, including wrist force sensing in a Universal Robots UR5e equipped with a Robotiq 2f-85 gripper, and finger force sensing using a Kinova Gen3 paired with an Openhand Model O gripper~\cite{odhner2014compliant}.

\section{Related Work}
\label{sec:RW}

Several previous methods exist which utilize force in the learning or control of robot movements. One of the most well-known examples is the extension of Dynamic Movement Primitives (DMPs)~\cite{SchaalDMP2005} sometimes known as Correlated DMPs or CorrDMP~\cite{calinon2010learning_CorrDMP}. In \cite{kormushev2011imitation}, a robot attempts to use CorrDMP to reproduce demonstrations with a specific given force profile. However, this method uses force profiles which were not given in the demonstration, and does not learn from force-inclusive demonstrations. Alternatively, \cite{pastor2011online} uses a version of DMP which replicates forces in reproductions as seen in previous demonstrations using a force control method. This assumes that the forces felt during demonstrations should also be experienced during reproductions, which may not always be the case. A method which does incorporate force in the learning process is \cite{gao2019learning}. This method models motions and forces using Gaussian Mixture Models (GMM) then finds reproductions using Gaussian Mixture Regression (GMR). This combination of GMM/GMR is also used in \cite{lin2012learning}, which focuses on learning motions and forces for grasping objects. Both these methods rely on a controller to reproduce the learned forces. Our method, on the other hand, incorporates forces directly into the reproduced trajectory.

Other methods have used force to learn motions. In \cite{rozo2013robot}, force and torque values are used to train Hidden Markov Models (HMMs), which are then used to determine a sequence of actions for the robot to follow using a modified version of GMR known as GMRa. This method learns forces well, and is shown to reproduce tasks better than demonstrators. However, this method lacks the ability to generalize to constraints such as via-point constraints. Our method both incorporates forces in the reproduction as well as any constraints given by the user. Additionally, multiple demonstrations are required to train the HMMs, whereas our method can work with either single or multiple demonstrations, making the process of demonstrating easier for users.

Other methods use forces learned in demonstrations to inform the control strategies used during reproduction. In \cite{lee2015learning}, a variable impedance control strategy is used to learn manipulation tasks with deformable objects. Alternatively, in \cite{shi2021combining}, a Reinforcement Learning (RL) agent is trained to find an optimal method for switching between position control and force of control of a robot during execution. This approach combines LfD with RL, in order to learn from demonstrations as well as learn by exploration. These methods, however, only use force in the control strategy employed by the robot. We simplify this process by incorporating force in the motion learning, allowing for any control strategy.

Reinforcement Learning has become an increasingly popular way for robots to learn new skills. Reinforcement Learning often requires many trials of interaction, which can be time-consuming or unsafe in the real-world. To alleviate this issue, sometimes training is performed in simulation, but this suffers from the transfer of knowledge from simulation to the real-world, which is not always accurate~\cite{kaspar2020sim2real}. Some methods combine LfD and RL, either by ``seeding'' the RL process with demonstrations to determine a reward function (known as Inverse Reinforcement Learning)~\cite{shiarlis2016inverse} or using RL to find optimal parameters for LfD representations~\cite{kober2014PoWER}. Reinforcement Learning techniques often suffer from the curse of dimensionality in the state space~\cite{sutton2018reinforcement}, so force is often excluded from RL methods. Additionally, reward functions or features are often hand-crafted, and force is not considered in these rewards. Our method can work with only a single demonstration, and does not require interaction time to find an optimal reproduction.

Many LfD representations attempt to learn only simple movements or primitives~\cite{pastorDMP2009}. However, in the real-world, many tasks are made up of a composition of these primitive motions. In order to break down longer tasks into motion primitives, segmentation must be performed. Segmentation algorithms detect changepoints throughout a demonstration, where the start and end of these changepoints corresponds to a motion primitive. Several methods exist for segmentation. In \cite{nakamura2017segmenting}, a segmentation algorithm using Gaussian Processes and Hidden Semi-Markov Models (GP-HSMM) is presented, which segments portions of data matching the same Gaussian Process. In \cite{song2020robot}, another method using a HSMM is used, which incorporates several modalities of robot trajectories, such as position, velocity, and torsion. Alternatively, \cite{carvalho2025parameter} proposes a segmentation method which can include multiple modalities, but uses representative demonstrations which must match the modalities of segmented demonstrations. The work of \cite{zhao2018fast} presents a multimodal segmentation algorithm, where the modalities used are video information and robot position. However, none of these segmentation methods explicitly include force as a modality, and none provide the means to include weighted modalities. Our segmentation method can incorporate any number of modalities, and allows for the weighting of these modalities to properly segment any given task.

\section{Methodology} 
\label{sec:method}

\subsection{Notation and Definitions}
We start with a given a set of demonstrations of a task $\mathcal{D} = \{\vect{D}^1, ..., \vect{D}^m\}$ with $m \geq 1$. The demonstrations can be collected through different methods such as kinesthetic teaching or teleoperation. An individual demonstration, denoted as $\vect{D}^i$, is a tuple of task-space trajectories (i.e., spatial data) as well as corresponding force data, $\vect{D}^i = ( X^i, \mathcal{F}^i )$. We denote the task-space trajectory as $X^i = [\vect{x}_1^i, \vect{x}_2^i, ..., \vect{x}_T^i]^{\top} \in \mathbb{R}^{T \times d}$, where $T$ is the number of observed $d$-dimensional points, and the force data as $\mathcal{F}^i = [\vect{f}_1^i, \vect{f}_2^i, ..., \vect{f}_T^i]^{\top} \in \mathbb{R}^{T \times d}$.   

\begin{figure}[ht]
    \centering
    \includegraphics[width=0.98\linewidth]{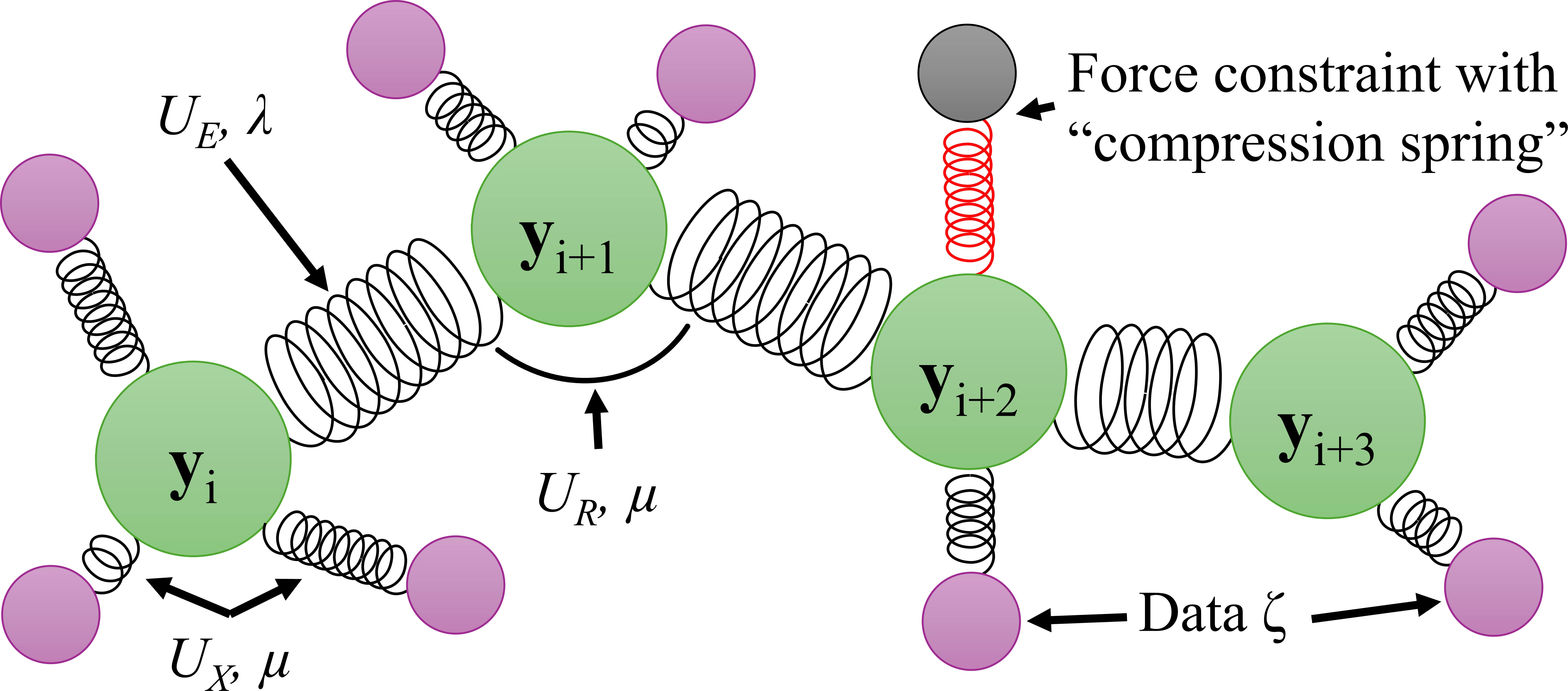}
    \caption{\small{Examples of elastic map data, nodes, and spring energies associated with approximation, stretching, and bending (details in Section~\ref{subsec:elmap})}}
    \label{fig:elmap-example}
\end{figure}

\subsection{Force-Constrained Elastic Maps}
\label{subsec:elmap}

In this work we present a method for force-constrained elastic maps, a multimodal Learning from Demonstration method which can learn robot trajectories based on not only the spatial properties of trajectories, but also the external force information. Given a set of one or more spatial trajectories, these are modeled using elastic maps with forces applied. Elastic maps seek to find an optimal trajectory via minimization of the following ``spring-like'' energies~\cite{gorban_zinovyev}:
\begin{enumerate*}[label=(\roman*)]
  \item the approximation energy $U_\mathcal{X}$ which pulls nodes to data,
  \item the stretching energy $U_E$ which pulls nodes together, and
  \item the bending energy $U_R$ which straightens adjacent edges.
\end{enumerate*}
The optimization of these energies results in a smooth reproduction which finds a mean approximation through the given trajectories~\cite{hertel2022ElMap}. The explicit formulation of the energies for an elastic map are:
\begin{align}
    U_\mathcal{X} &= \frac{1}{\sum_{\vect{\zeta}_j} w_j} \sum_{i = 1}^{N} \sum_{\vect{\zeta}_j \in \kappa_i} w_j || \vect{\zeta}_j - \vect{y}_i ||_2^2, \label{eq:U_X1}\\
    U_E &= \lambda \sum_{i = 1}^{N-1} || \vect{y}_{i+1} - \vect{y}_i ||_2^2,  \label{eq:U_E1}\\
    U_R &= \mu \sum_{i = 1}^{N-2} || \vect{y}_i - 2 \vect{y}_{i+1} + \vect{y}_{i+2} ||_2^2, \label{eq:U_R1}
\end{align}
where $\vect{\zeta} = [\vect{\zeta}_1, \vect{\zeta}_2, ..., \vect{\zeta}_M]$ is the stacked demonstration data (i.e., $\vect{\zeta} = [X^1, X^2, ..., X^m]^{\top}$), $w_j$ is the weight of data $\vect{\zeta}_j$, $\vect{y} = [\vect{y}_1, \vect{y}_2, ..., \vect{y}_N]$ is the nodes in the elastic map, $\kappa_i$ is the cluster of data for node $\vect{y}_i$, $||\cdot||_n$ is the $L^n$-norm, and $\lambda$ and $\mu$ are the stretching and bending constants, respectively. This formulation can handle spatial data, as the approximation energy pulls the nodes towards the given data in space. In this paper, however, we introduce a novel method for handling force constraints. Since the approximation energy models connections between data and nodes as springs, we introduce a ``spring-like'' force to be included in the data and optimization (see Fig.~\ref{fig:elmap-example}).

To include a force as a data point in the elastic map, the energy of the spring connection must be related to the force. The force exerted by a spring is given as 
\begin{equation}
    f = -k\Delta
\end{equation}
according to Hooke's law, where $f$ is the force, $k$ is the spring constant, and $\Delta$ is the displacement of the spring from equilibrium. The antiderivative of this force equation is 
\begin{equation}
    u = -\frac{1}{2}k\Delta^2,
\end{equation}
where $u$ is the energy of the spring. The energy of this force-derived spring must match the energy in \eqref{eq:U_X1}. Since $f$ is known from the demonstration and $\Delta$ is given by the distance from where the force was applied to the current placement of a node, we can find $k$ via $k = -\frac{f}{\Delta}$. The $k$ value corresponds with the weight $w$ of a data point, so the force can be included by inserting a data point into $\vect{\zeta}$ at the position (perturbed by a small $\delta$) the force was felt, with weight $-\frac{f}{\Delta}$. However, the weight $w$ in \eqref{eq:U_X1} is a scalar value whereas the force felt from demonstrations $\vect{f}$ is a $d$-dimensional vector value. Therefore, all weights must be vectors for consistency. The weight for a standard 3-D point along the demonstration trajectory may be $\vect{w} = [1, 1, 1]$, which pulls nodes towards the data in all directions. The weight for a force data point would be $[-\frac{f_x}{\Delta_x}, -\frac{f_y}{\Delta_y}, -\frac{f_z}{\Delta_z}]$, which pushes nodes away according to the calculated spring constant $k$. Note that since weights are now a vector, we must change the approximation energy to
\begin{equation}
    U_\mathcal{X} = \frac{1}{\sum_{\vect{\zeta}_j} ||\vect{w}_j||_2} \sum_{i = 1}^{N} \sum_{\vect{\zeta}_j \in k_i} || \vect{w}_j \odot \vect{\zeta}_j - \vect{w}_j \odot \vect{y}_i ||_2^2, \label{eq:U_X2}\\
\end{equation}
where $\odot$ is the Hadamard product. Also note that the small displacement $\delta$ of constraint nodes is needed such that initially $\Delta$ is non-zero. This informs the approximation energy such that it avoids approximating forces felt during the demonstrations; the larger the force, the stronger the avoidance.

To find an optimal placement of nodes for an elastic map, an Expectation-Maximization algorithm is used~\cite{hertel2022ElMap}. In the Expectation step, each data point is clustered to the closest node. In the Maximization step, the nodes are moved such as to minimize the total energy of the current clustering as 
\begin{equation}
    \underset{\vect{y}}{\text{minimize }} g(\vect{y}) = U_\mathcal{X} + U_E + U_R \numberthis\label{eq:opt}
\end{equation}
which results in an optimal map for that clustering. This process is then repeated, with a new clustering computed on the updated node positions and that placement energy minimized, until convergence (clusterings are unchanged). 

\subsection{Convex Formulation of Energies}

In order to quickly minimize the energies presented, we present a convex formulation of the elastic map energies. To achieve this, we start with a presentation of the convex form of equations \eqref{eq:U_X1}-\eqref{eq:U_R1}:
\begin{align}
    U_\mathcal{X} &= \frac{1}{\sum_{\vect{\zeta}_j} w_j} || \vect{A}\vect{y} -\vect{B}\vect{\zeta} ||_2^2 \label{eq:U_Y_new}\\
    U_E &= \lambda \vect{y}^\top\vect{E}^\top\vect{E}\vect{y} \label{eq:U_E_new}\\
    U_R &= \mu \vect{y}^\top\vect{R}^\top\vect{R}\vect{y}, \label{eq:U_R_new}
\end{align}
where $\vect{A}$ and $\vect{B}$ are weighting and clustering matrices, derived from the clustering process. The clustering matrix $\vect{B}$ is defined as
\begin{equation}
    \vect{B}_{ij} = 
    \begin{cases} 
      w_j & \vect{\zeta}_j \in \kappa_i,\\
      0 & \text{otherwise},
   \end{cases}   
\end{equation}
which assigns a weight to $\vect{B}_{ij}$ if data $\vect{\zeta}_j$ is clustered to node $\vect{y}_i$. $\vect{A}$ is a diagonal weighting matrix such that 
\begin{equation}
    \vect{A}_{ii} = \sum_j \vect{B}_{ij}
\end{equation}
which weights all nodes appropriately according to \eqref{eq:U_X1}. The edge matrix $\vect{E}$ and rib matrix $\vect{R}$ are defined as:
\begin{equation}
\vect{E} = 
\left [ \begin{smallmatrix}
%\begin{bmatrix}
 -1 & 1 &  0 & \cdots & 0 \\
0 &  -1 & 1 & \cdots & 0 \\
\vdots & \ddots &\ddots & \ddots & \vdots \\
0 & \cdots & 0 & -1 & 1   \\
%\end{bmatrix}, \nonumber 
%\end{equation}
\end{smallmatrix} \right ],
%\begin{equation}
\vect{R} = 
\left [ \begin{smallmatrix}
%\begin{bmatrix}
1 & -2 &  1 & 0 & \cdots & 0 \\
0 & 1 & -2 & 1 & \cdots & 0 \\
\vdots & \ddots & \ddots & \ddots & \ddots & \vdots\\
0 & \cdots & 0 & 1 & -2 & 1   \\
%\end{bmatrix}. \nonumber
\end{smallmatrix} \right ], \nonumber
\end{equation}
which correspond to the differences taken in \eqref{eq:U_E1} and \eqref{eq:U_R1}, respectively~\cite{hertel2023confidence}. The convex formulation of the approximation energy in \eqref{eq:U_Y_new} uses scalar weights and therefore cannot be used with force-inclusive demonstrations. In order to use vector weights, we first alter the data $\vect{\zeta}$ and nodes $\vect{y}$ such that they are 1-dimensional column vectors, stacking each dimension onto another. For example, this results in a change such that the nodes go from dimension $\mathbb{R}^{N \times d}$ to  $\mathbb{R}^{Nd \times 1}$. We denote the stacked data and stacked nodes as $\vect{\xi}$ and $\vect{\gamma}$, respectively. Originally, $\vect{B}$ also uses scalar weights, and cannot incorporate forces in directional weights. Therefore, we use a different clustering matrix $\vect{C}$ where if $\vect{\zeta}_j \in \kappa_i$, then $\vect{C}_{i+qN, j+qM} = \vect{w}_{j, q+1}$ for $q = {0, 1, ..., d-1}$. We also introduce a new diagonal weighting matrix $\vect{W}$, where $\vect{W}_{ii} = \sum_j \vect{C}_{ij}$. Altogether, this results in the convex form of \eqref{eq:U_X2} as
\begin{equation}
    U_\mathcal{X} = \frac{1}{\sum_{\vect{\zeta}_j} ||\vect{w}_j||_2} || \vect{W}\vect{\gamma} -\vect{C}\vect{\xi} ||_2^2,\label{eq:U_X2_new}
\end{equation}
which together with \eqref{eq:U_E_new} and \eqref{eq:U_R_new} can be used to find the optimal node placement of a force-constrained elastic map. These nodes can then be used as a trajectory for a robot manipulator to follow.

\subsection{Force-Based Segmentation}

Segmentation is often done solely with spatial data~\cite{nakamura2017segmenting}, and rarely with other modalities~\cite{hertel2024reusable_skills}. Many robotic tasks, however, rely inherently on force data (e.g., ironing~\cite{kormushev2011imitation}). Force can also be used as an additional modality to derive context from demonstrations. For example, when picking an object, an increase in force can indicate contact. To detect context in force-inclusive demonstrations, we propose a novel method for weighted probabilistic multimodal segmentation. Our segmentation process takes multiple data streams, finds keypoints for each individual data stream, then probabilistically combines these keypoints to find changepoints which segment the overall demonstration.

First we find the keypoints of data streams within a given demonstration $\vect{D}$ containing multiple data streams, with an individual data stream denoted $\vect{S}_i$, where $i$ indexes the  data stream. To find the keypoints of $\vect{S}_i$, a thresholding window algorithm is performed on the third derivative of the data stream $\vect{S}_i'''$~\cite{hertel2024reusable_skills}. A peak in the third derivative of a signal (e.g., jerk for spatial data) indicates a change in the direction of the curvature of the signal. These changes or shifts in direction are present in other derivatives as well, but the third derivative shows repeatable patterns of these changes. However, noise in the original data stream translates to large noise when taking the derivative, therefore a moving average of the third derivative is used. If the third derivative maintains a value above a certain threshold for a certain window, the start of that window is considered a keypoint. New keypoints cannot be created until the third derivative drops below the threshold, and then returns above it. This results in a list of keypoints for that data stream, but different data streams may not concur with their keypoints. For example, take gripper position and arm position as different data streams. The arm only moves while the gripper is still, and the gripper only closes when the arm remains fixed. Therefore, these data streams would have very different keypoints. To combine the keypoints across different data streams, we propose a weighted probabilistic combination method. First, keypoints for a specific data stream are modeled as probabilities of changepoints, where a keypoint corresponds to a Gaussian probability of a changepoint $\mathcal{N}(t, \sigma^2)$, where $t$ is the timestep of the keypoint and $\sigma^2$ is the variance based on the size of the moving window average. This creates a probability mass function for each data stream, giving the probability that a specific timestep is a changepoint. The probability mass function of data stream $\vect{S}_i$ is denoted $p_\eta^{\vect{S}_i}(t)$, where $\eta$ represents a changepoint. The probability mass functions are combined into the probability mass function of changepoints over the entire demonstration $p_\eta^{\vect{D}}(t)$ as
\begin{equation}
    p_\eta^{\vect{D}}(t) = \prod_i v_i p_\eta^{\vect{S}_i}(t), \label{eq:prob-comb}
\end{equation}
\noindent where $v_i$ indicates the importance of changepoints detected in data stream $\vect{S}_i$. These importances are task-dependent and tuned manually, for instance, in force-constrained tasks such as ironing, the force data streams are more important, but in some position-based tasks such as writing, the position is more important. Alternatively, data streams from other modalities (e.g., image data using optical flow~\cite{horn1981determining}) could be measured and incorporated depending upon the task. Finally, we sample from the changepoint probability distribution. If a given window contains more samples than some threshold, that is considered a consensus and a changepoint is placed at the mean of those samples. Using these changepoints we segment the demonstration into parts that represent primitive movements, which can be learned using force-constrained elastic maps explained in the previous section.

\begin{figure}[t]
    \centering
    \includegraphics[width=0.98\linewidth]{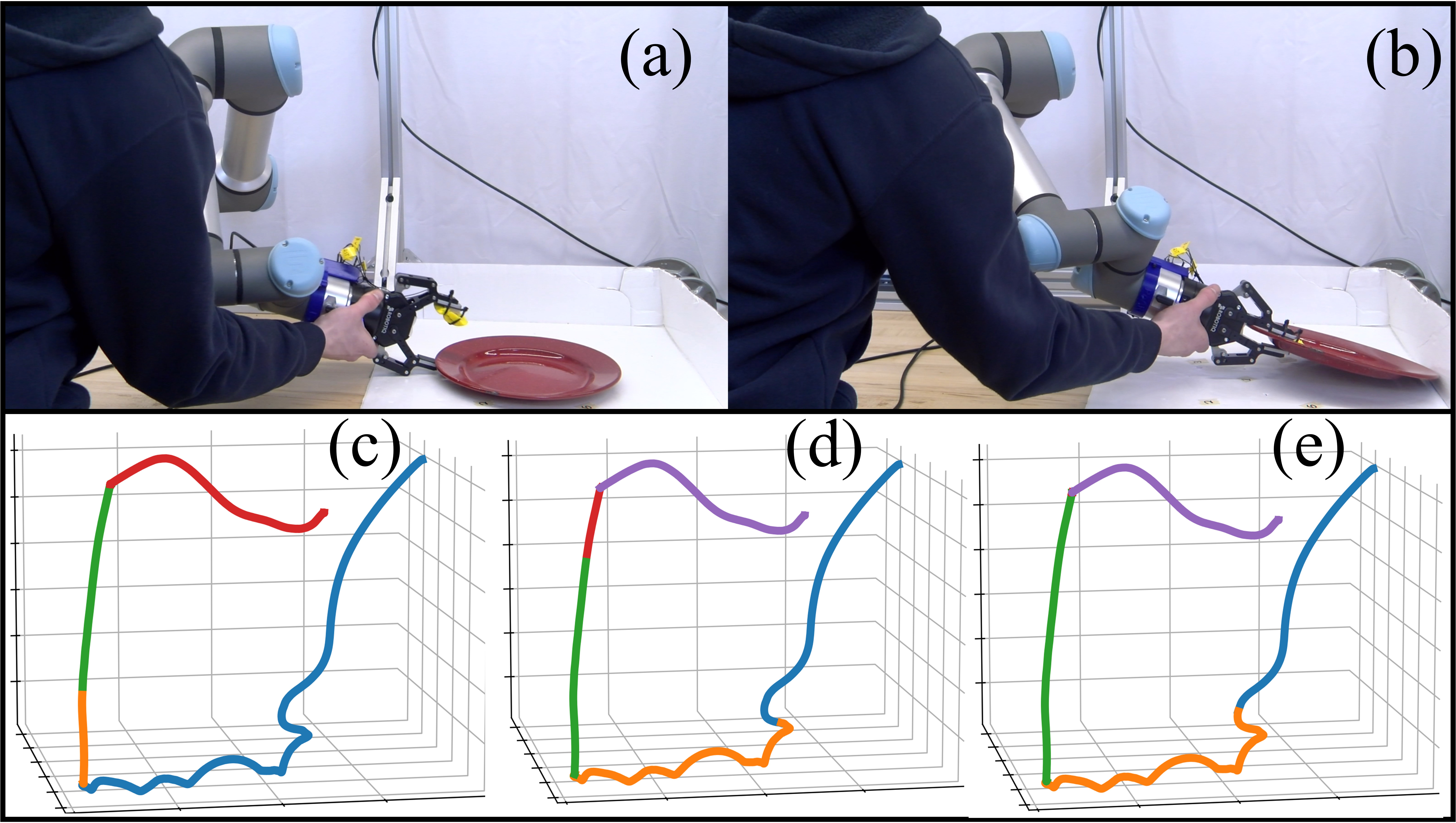}
    \caption{\small{(a-b) Example of picking up a plate by sliding it against a wall. Results for position-only segmentation (c), uniform multimodal segmentation (d), and multimodal segmentation with custom weights (e). Each colored portion of the trajectories in (c), (d), and (e) represents a segment. The custom weights find a more intuitive segmentation of the motion, with clear segments for the approach, slide, and pick movements. Both (c) and (d) segment the approach movement into two sections, whereas (e) correctly keeps this movement whole.}}
    \label{fig:segmentation-compare}
\end{figure}

\begin{figure*}[t]
    \centering
    \includegraphics[width=0.95\linewidth]{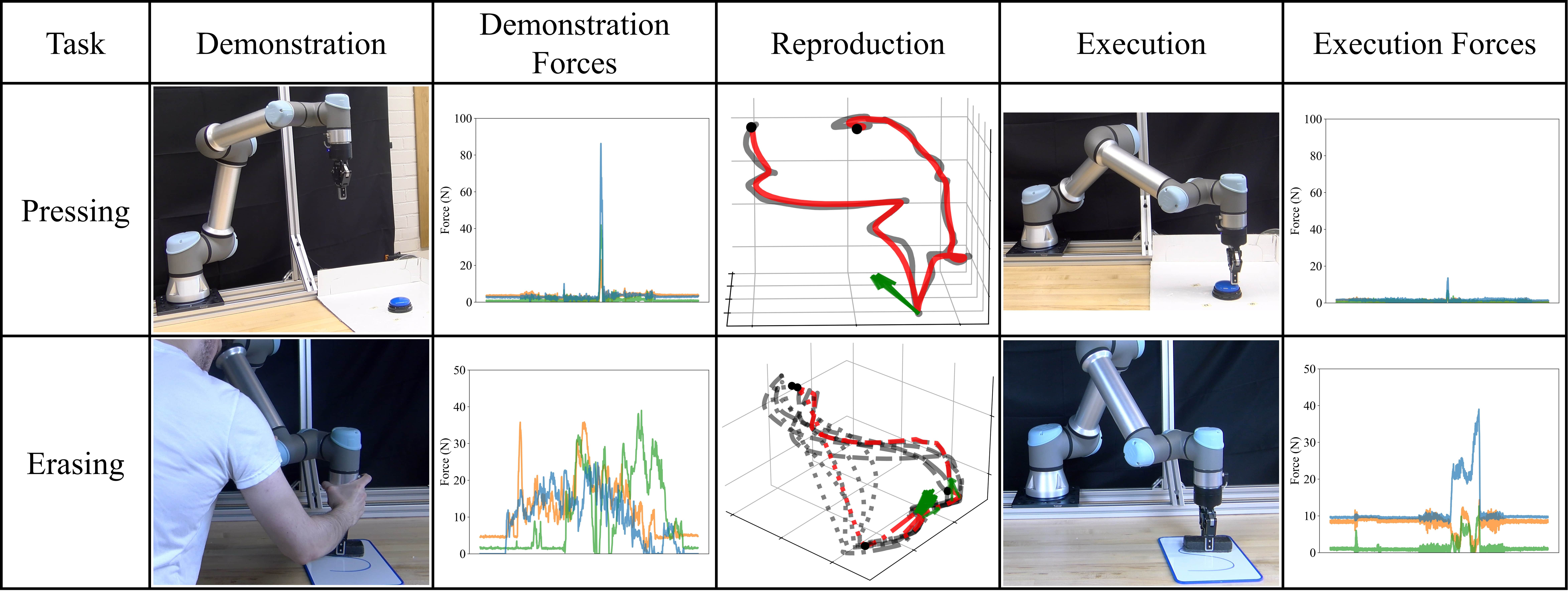}
    \includegraphics[width=0.8\linewidth]{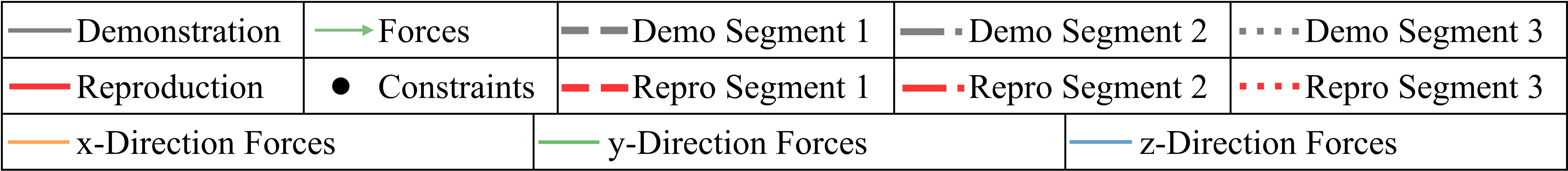}
    \caption{\small{Several tasks performed using a UR5e manipulator. In each task type, a demonstration is given, and a reproduction using force-constrained LfD is found. The reproductions successfully complete the task while avoiding unnecessary forces.}}
    \label{fig:ur5e-exp}
\end{figure*}

\section{Experiments}
\label{sec:exps}

We validate our method across several environments. First, we apply our segmentation to a dataset including the spatial and force modalities, comparing the efficacy of force-important weighting. Next, we validate our learning framework across several tasks, including pressing and erasing.\footnote{Accompanying video: \url{https://youtu.be/_0xq8XFP-PQ}} In our experiments, we use two robotic platforms---a Universal Robots UR5e with a Robotiq 2f-85 gripper, and a Kinova Gen3 with an Openhand Model O gripper. While the UR5e setup is equipped with a wrist force/torque sensor, the Kinova setup relies on fingertip force sensing. In all experiments, reproductions are computed \textit{a priori} then followed by a low-level controller.\footnote{Details for implementation are available at: \\ \url{https://github.com/brenhertel/Force-Constrained-ElasticMaps} }

\begin{table}[h]
\centering
\caption{\small Weights ($v_i$ in Eq. \eqref{eq:prob-comb}) of different modalities used for segmentation of grasping dataset. Results of segmentation are shown in Table~\ref{tab:segments-compare}.}
\label{tab:segments-weights}
%\setlength\tabcolsep{3.3pt}
%\scriptsize
\begin{tabular}{cccccc} \toprule
 &
  \begin{tabular}[c]{@{}c@{}}Cartesian\\ Trajectory\end{tabular} &
  \begin{tabular}[c]{@{}c@{}}Joint\\ Trajectory\end{tabular} &
  \begin{tabular}[c]{@{}c@{}}Gripper\\ Position\end{tabular} &
  \begin{tabular}[c]{@{}c@{}}Wrist\\ Force\end{tabular} &
  \begin{tabular}[c]{@{}c@{}}Finger\\ Force\end{tabular} \\ \midrule
\begin{tabular}[c]{@{}c@{}}Position\\ Only\end{tabular}   & 0.50 & 0.50 & 0.00 & 0.00 & 0.00 \\
\begin{tabular}[c]{@{}c@{}}Uniform\\ Weights\end{tabular} & 0.20 & 0.20 & 0.20 & 0.20 & 0.20 \\
\begin{tabular}[c]{@{}c@{}}Custom\\ Weights\end{tabular}  & 0.10 & 0.10 & 0.10 & 0.35 & 0.35  \\
\bottomrule
\end{tabular}
\end{table}

\begin{table}[h]
\centering
\caption{\small Results for number of segments of performing segmentation across a grasping dataset, averaged over 5 runs, with 60 demonstrations in each run. Custom weights incorporate more force in the segmentation, and result in better segmentation overall (finding more than 4 segments would result in over-segmentation). Visual segmentation results are shown in Fig.~\ref{fig:segmentation-compare}.}
\label{tab:segments-compare}
%\setlength\tabcolsep{3.3pt}
%\scriptsize
\begin{tabular}{@{}lccc@{}}\toprule
& Position Only & Uniform Weights & Custom Weights \\ \midrule
Mean & 4.09 & 4.11 & 3.86  \\
Std. Dev. & 0.95 & 0.87 & 0.84  \\
\bottomrule
\end{tabular}
\end{table}

\subsection{Segmentation of Force-Constrained Demonstrations}

We collect a dataset of force-inclusive demonstrations where demonstrators grasp several objects across varying locations.\footnote{ Dataset: \url{https://github.com/brenhertel/GBTI-Grasping-Dataset}} This dataset consists of 3 objects (cube, mug, and plate), 4 demonstrators, and 5 object locations, for a total of 60 demonstrations. These demonstrations contain information on the joint positions/velocities/efforts, end-effector position/orientation, wrist force/torque, and gripper position/force. Demonstrations were captured with a Universal Robots UR5e with a Robotiq 2f-85 gripper, modified with a force sensor on one finger. We compare the segmentation of these demonstrations with and without weighting the importance of force in the task. Additionally, we present a baseline of position-only segmentation. The weights for these segmentations can be seen in Table~\ref{tab:segments-weights}. Qualitatively, we are looking for about 3-4 segments for a grasp operation: the approach, grasp, and retract. Sometimes, however, the grasp can be decomposed into an alignment or slight push, then the grasp. More than 4 segments would be considered an over-segmentation of the demonstration. The segmentation results are shown in Table~\ref{tab:segments-compare}, with an example comparing two segmentations in Fig.~\ref{fig:segmentation-compare}. As can be seen, an increase in the importance of the force modality resulted in a more accurate segmentation, with segments aligning with the expectation of the motion, on average staying within the 3-4 segment range ($p<0.01$ for a significantly lower average according to Student's t-Test) and with less variance. It is often seen that position-only and uniform weighting create small erroneous segments, or combine multiple portions of a task which should be segmented. Segmentation can be used to detect the approach, pick, and retreat motions, which reveal the context of the demonstration.

\subsection{One-Shot Learning from Wrist Force-Constrained Demonstrations}

After evaluating the capabilities of the segmentation process, we validate the proposed LfD framework in various real-world scenarios. We first test our method using a Universal Robots UR5e with a Robotiq 2f-85 gripper (without finger force sensing, unlike the previous section). 

\noindent \textbf{Pressing:} First, we validate our method using a simple button pressing task. As seen in Fig.~\ref{fig:ur5e-exp}, in this task, the button is placed on the table, and the robot is guided to press the button using teleoperation through Virtual Reality (VR) with a Meta Quest 2~\cite{boguslavskii2023shared}. This teleoperation method does not include any haptic feedback unlike, for example, kinesthetic teaching would, and therefore operators cannot accurately control the force applied to the button. In some demonstrations, this caused excessive forces at the pressing point which should be avoided in reproductions (demonstrations that triggered the robot's safety limits and caused an emergency stop were excluded). Our method, however, learns to avoid high-magnitude forces, while still finding an accurate reproduction of the task. The demonstration and reproduction of this pressing task can be seen in Fig.~\ref{fig:ur5e-exp}, along with the associated forces of the demonstration and reproduction. The change in force of the reproduction compared to the demonstration can be seen in Table~\ref{tab:forces-compare}. In this task, the maximum force sees an 84.35\% reduction in magnitude (a reduction of over 70 N), greatly decreasing the amount of force applied to the robot. Despite the drastic force reduction during the reproduction, the robot was able to perform the task successfully. 

\noindent \textbf{Erasing:} We also perform an erasing task using the UR5e. In this task, the letter ``S'' was drawn on a whiteboard. The robot was  taught kinesthetically through 5 demonstrations to approach the whiteboard, follow the path of the letter to erase it, then retreat. The kinesthetic teaching of this task allowed the demonstrator to have better control over the forces applied to the robot (unlike the pressing task, no safety limits were triggered), but on average results in more force than is necessary to complete the task. Note that stable forces are required for this task, but exact imitation is not necessarily needed. We performed our force-based segmentation, which correctly identifies the approaching, erasing, and retreating across all 5 demonstrations (see Fig.~\ref{fig:ur5e-exp}). A slight increase is seen in the maximum force experienced by the robot as shown in Table~\ref{tab:forces-compare}, however, the average force felt across the reproduction is reduced by almost a third compared to the demonstrations. This shows that our method is able to reproduce different force profiles compared to the demonstrations, but still apply forces necessary to complete tasks. Additionally, we generalize to novel start and end-points for each segment, showcasing the ability to meet novel constraints.

\begin{table}[h]
\centering
\caption{\small Changes in the maximum and average forces experienced during demonstrations compared to reproductions for the tasks shown in Fig.~\ref{fig:ur5e-exp}. }
\label{tab:forces-compare}
%\setlength\tabcolsep{3.3pt}
%\scriptsize
\begin{tabular}{@{}lrr@{}}\toprule
& Pressing & Erasing \\ \midrule
Max Force Change \%   & -84.35\% & 0.04\%  \\
Avg. Force Change \%  & -66.64\% &  -32.11\%  \\
\bottomrule
\end{tabular}
\end{table}

\begin{figure}[t]
    \centering
    \includegraphics[width=0.98\linewidth]{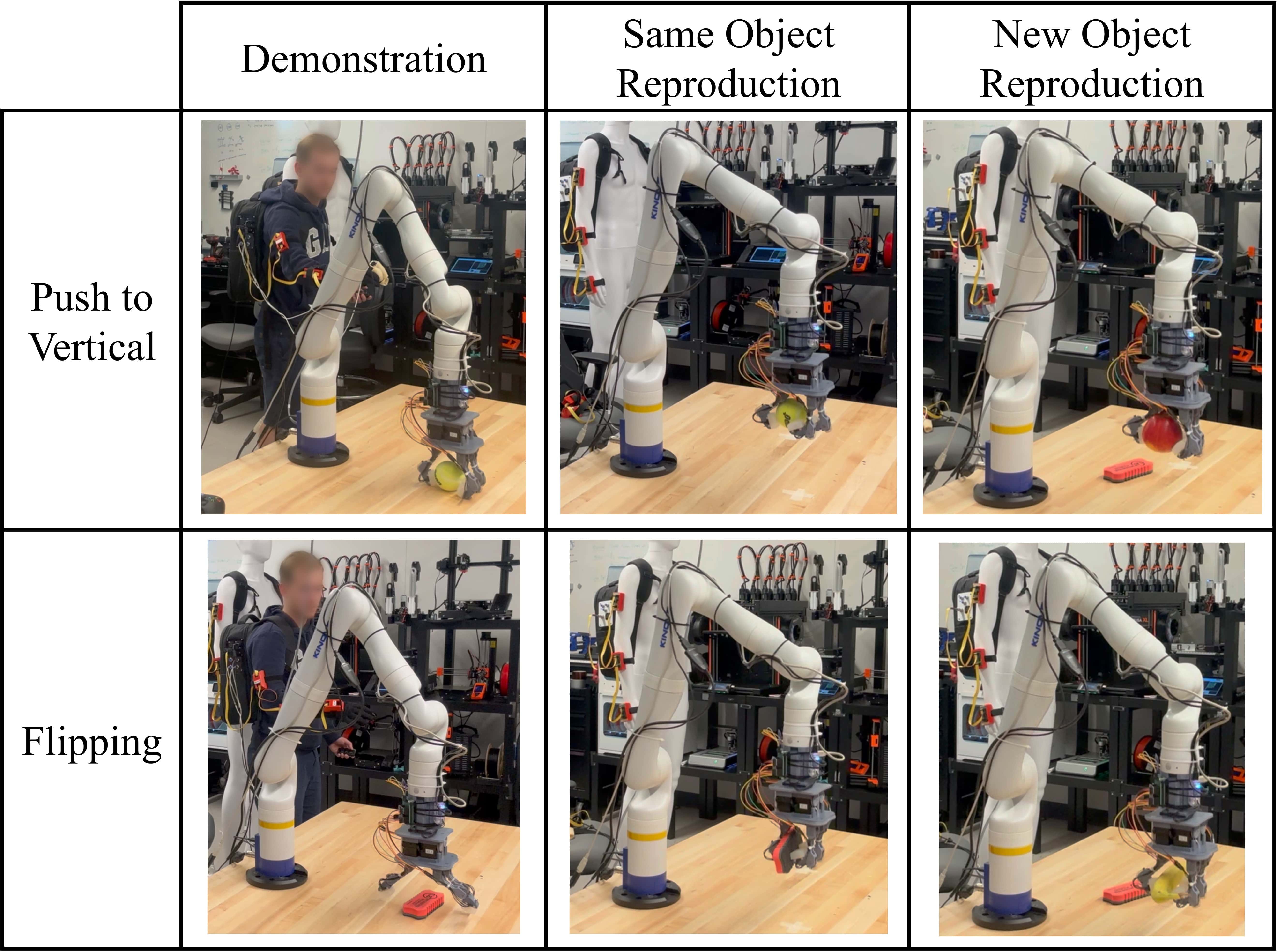}
    \caption{\small{Demonstrations and reproductions across several different objects for a grasping task. A grasp is demonstrated using either the ``push to vertical'' or ``flipping'' methods. This grasp is then learned using force-inclusive demonstrations and reproduced across several objects, including new objects unknown to the robot (an apple and a banana), which it is still able to grasp successfully.}}
    \label{fig:grasping-exp}
\end{figure}

\subsection{One-Shot Learning from Finger Force-Constrained Demonstrations}

Different robots have different sensors, and a method which uses force should be able to learn with any available force sensing. To validate our framework on a different sensor configuration, we used a Kinova Gen3 equipped with an Openhand Model O gripper, with the gripper modified to have force sensing in each of the three fingertips. A demonstration is given using a teleoperation device with haptic feedback as described in~\cite{thakur2024tetherless}. During the demonstration, the user teleoperated the robot to move down, grasp an object, then retract the arm. We used two different objects with two different grasping techniques, a ``push to vertical'' and a ``flipping'' motion. These demonstrations were separated into two time-aligned sub-demonstrations: a demonstration for the arm and a demonstration for the gripper. Each of these were learned in parallel using our proposed LfD representation. To reproduce these skills, we executed both learned trajectories simultaneously. We reproduced the grasping with the same object as well as several different objects as seen in Fig.~\ref{fig:grasping-exp}. The results highlight the capabilities of the proposed framework to generalize across different sensor configurations and tasks.

\begin{figure}[ht]
    \centering
    \includegraphics[width=0.98\linewidth]{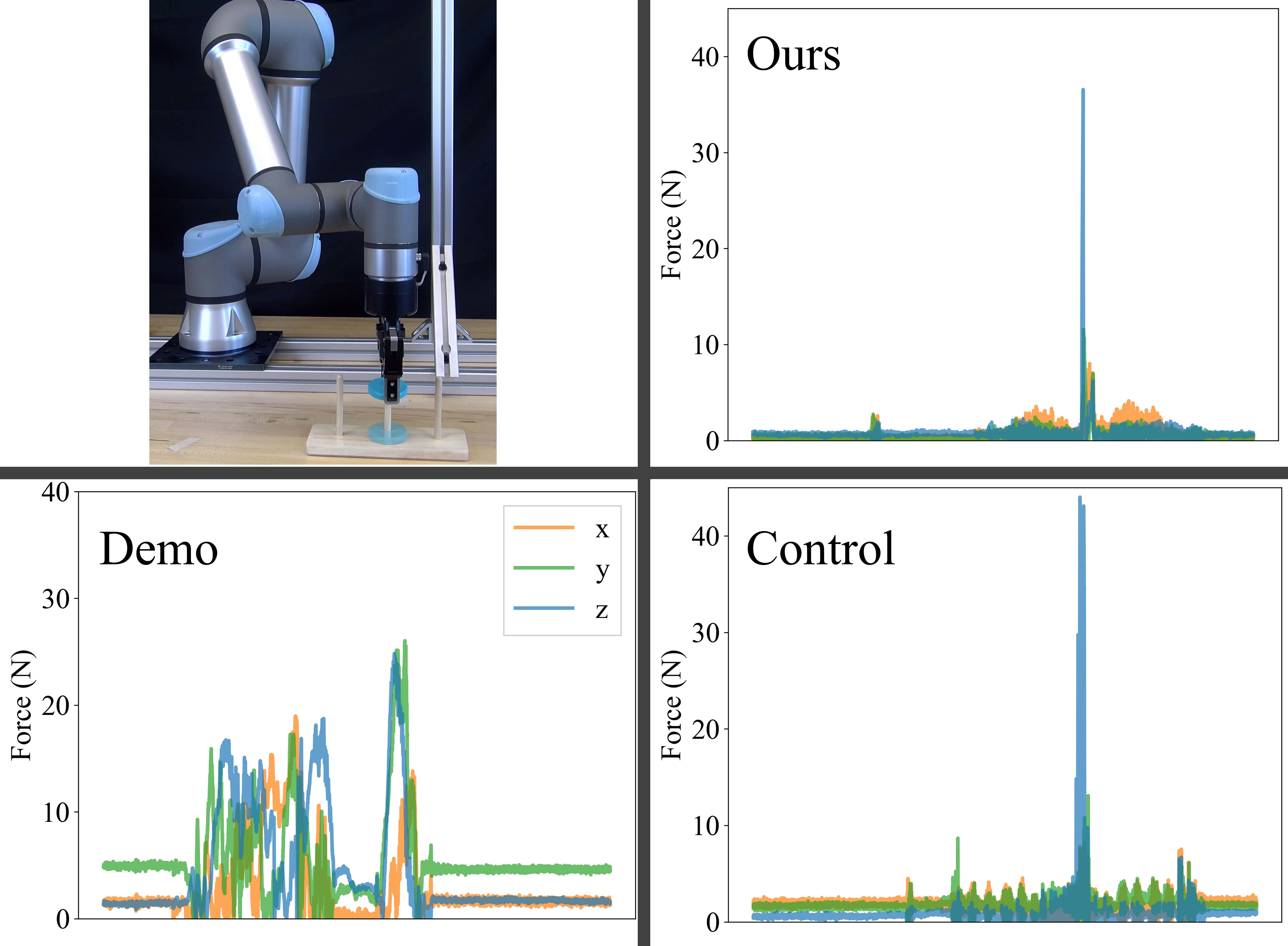}
    \caption{\small{Comparison of forces between a demonstration of a peg-in-hole task and reproductions using force control and our method. For quantitative results see Table.~\ref{tab:comp-exp}.}}
    \label{fig:comp-exp}
\end{figure}

\begin{table}[h]
\centering
\caption{\small Changes in the maximum (Max) and average (Avg.) forces (N) for each dimension experienced during demonstrations compared to reproductions for the tasks shown in Fig.~\ref{fig:comp-exp}. }
\label{tab:comp-exp}
%\setlength\tabcolsep{3.3pt}
%\scriptsize
\begin{tabular}{@{}lrrrrrr@{}}\toprule
Method & X Max & Y Max & Z Max & X Avg. & Y Avg. & Z Avg. \\ \midrule
Demo & 18.96 & 26.02 & \textbf{24.82} & 2.91 & 5.49 & 4.52 \\
Control & 8.19 & 13.08 & 44.01 & 1.73 & 1.83 & 1.03 \\
Ours & \textbf{8.00} & \textbf{11.57} & 36.55 & \textbf{0.54} & \textbf{0.50} & \textbf{0.76} \\
\bottomrule
\end{tabular}
\end{table}

\subsection{Comparison Against Force-Based Control}

We additionally compare our method which learns from force-constrained demonstrations against executions using basic hybrid force/position control~\cite{raibert1981hybrid} in a peg-in-hole task. In this task, the robot begins with a disc in its gripper. This disc has a small hole which must be placed on a peg as seen in the top right of Fig.~\ref{fig:comp-exp}. Only wrist forces are used in this experiment, and only a single demonstration is provided. The hybrid force/position control was set up such that the robot followed the demonstration according to position control until it reached a certain threshold of forces felt, at which point force control took over. Once forces returned below that threshold, the execution once again followed position control. Our method was set to use the same threshold to preempt the reproduction to avoid those forces. The forces felt in this experiment for the demonstration, force control, and our method are shown in Fig.~\ref{fig:comp-exp}, with maximum and average forces for each direction shown in Table~\ref{tab:comp-exp}. It can be seen that for nearly all directions, both the maximum and average forces for our method are the lowest, with the exception of the maximum in the Z direction. We believe that for both the force control reproduction and reproduction using our method, an error in our vision system used to detect the center of the peg caused the robot to be slightly off (but not so off that it was unable to complete the task), resulting in a slight bump when first approaching the peg. The vision portion of this task is not the focus of this paper, although this stresses the importance of having accurate visual comprehension of a task. However, if we look at the forces, it can be seen that our method is able to avoid excessive forces throughout the task, and even perform better in the case of unexpected force perturbations caused by errors from other systems.

\section{Conclusions \& Future Work}
\label{sec:conc}

We have proposed a novel one-shot multimodal Learning from Demonstration framework for the segmentation, encoding, and reproduction of force-constrained tasks. Our framework employs a multimodal segmentation technique to assess the importance of different modalities over time. We also incorporated force constraints into elastic maps, allowing the encoding and modeling of force-constrained tasks. We validate the effectiveness of our framework on five real-world tasks, demonstrating its versatility across various force-sensing platforms. Our results have shown a reduction in the forces experienced by the robot, resulting in safer execution.

In this work we focused on incorporating force constraints into the task learning process. Integrating this approach with existing force control methods is a promising next step, especially if executed force profiles should be more similar to those demonstrated. Our method currently only uses forces to learn motions prior to execution, but incorporating force during execution as well could lead to safer execution and more robust generalization. Additionally, our results suggest that the proposed method performs better in some tasks than others. Further research could explore its effectiveness across a broader range of tasks, potentially refining the approach through task-specific tuning. We occasionally see higher force spikes using our method, which could likely be addressed in the motion generation using some constraints. A method which could automatically find and apply these constraints to further reduce unnecessary forces could increase the efficacy of this approach. Additionally, the weights for our multimodal segmentation method are user-defined. In the future, automatically determining these weights could reduce the burden on users.
    
\section*{Acknowledgements}

This research is supported in part by the National Science Foundation (FRR-2237463) and Amazon Robotics through the Greater Boston Tech Initiative (GBTI) program.

The authors would like to thank Meriem Elkoudi, Wendy Carvalho, Owen O'Brien, and Ryan Donald from the PeARL Lab at the University of Massachusetts Lowell; Shilpa Thakur from the Soft Robotics Lab at Worcester Polytechnic Institute; and Nikita Boguslavskii from the HiRo Lab at Worcester Polytechnic Institute for their help with experiments.

\typeout{}
\bibliographystyle{IEEEtran}
\bibliography{references}

\end{document}